\pdfoutput=1
\documentclass[runningheads]{llncs}

\usepackage[T1]{fontenc}
\usepackage[utf8]{inputenc}
\usepackage{graphicx}
\usepackage{xcolor}
\usepackage{booktabs}
\usepackage{multirow}
\usepackage{makecell}
\usepackage{array}
\usepackage{microtype}
\usepackage{xurl}
\usepackage[pagebackref=true,breaklinks=true]{hyperref}
\hypersetup{
colorlinks,
linkcolor={red!75!black},
citecolor={blue!75!black},
urlcolor={blue!75!black},
}
\usepackage{cleveref}
\usepackage{csquotes}
\usepackage{amssymb}
\usepackage[inline]{enumitem}
\usepackage[nolist]{acronym}
\usepackage{listings}

\lstdefinelanguage{json}{%
  basicstyle=\ttfamily\scriptsize,
  showstringspaces=false,
  breaklines=true,
  numbers=none,
  columns=fullflexible,
  morestring=[b]",
  stringstyle=\color{teal!70!black},
  keywordstyle=\color{blue!70!black},
  commentstyle=\color{gray},
  literate=
    {ä}{{\"a}}1 {ö}{{\"o}}1 {ü}{{\"u}}1
    {Ä}{{\"A}}1 {Ö}{{\"O}}1 {Ü}{{\"U}}1
    {ß}{{\ss}}1 {é}{{\'e}}1
}
\newcommand{\figplaceholder}[2][.9\linewidth]{%
  \IfFileExists{#2}{\includegraphics[width=#1]{#2}}{%
    \fbox{\parbox[c][3.0cm][c]{#1}{\centering}}%
  }%
}

\newlength{\smalltablefontsize}
\newlength{\smalltableblskip}

\newcommand{\smalltablefont}{%
  \fontsize{\smalltablefontsize}{\smalltableblskip}\selectfont
}

\begin{document}

\title{From Pixels to Structure: Lightweight Vision-Language Models for Document OCR and Structured JSON Extraction}
\titlerunning{Title Suppressed Due to Excessive Length}

\author{Uddipan Basu Bir\inst{1} \and
Vincent Christlein\inst{1} \and
Andreas Maier\inst{1} \and
Mathias Zinnen\inst{1}}
\authorrunning{U. Basu Bir et al.}
\institute{Friedrich-Alexander-Universit\"at Erlangen--N\"urnberg, Germany\\
\email{\{uddipan.bir.basu,vincent.christlein,andreas.maier,mathias.zinnen\}@fau.de}}

\maketitle

\acrodef{OCR}{Optical Character Recognition}
\acrodef{VLM}{Vision-Language Model}
\acrodef{JSON}{JavaScript Object Notation}
\acrodef{CER}{Character Error Rate}
\acrodef{ANLS}{Approximate Normalized Levenshtein Similarity}
\acrodef{mAP-F1}{mean Average Precision F1}
\acrodef{KIE}{Key Information Extraction}
\acrodef{GLAM}{Galleries, Libraries, Archives, and Museums}
\acrodef{API}{Application Programming Interface}
\acrodef{GPU}{Graphics Processing Unit}
\acrodef{VRAM}{Video Random-Access Memory}
\acrodef{LoRA}{Low-Rank Adaptation}
\acrodef{QLoRA}{Quantized Low-Rank Adaptation}
\acrodef{TPE}{Tree-structured Parzen Estimator}
\acrodef{CLAHE}{Contrast Limited Adaptive Histogram Equalization}

\begin{abstract}
While massive, closed-source \acp{VLM} set strong benchmarks for document understanding, their dependence on commercial \acp{API} limits adoption in institutional archives due to data autonomy concerns, recurring costs, and the environmental footprint of hyperscale computing. This is especially acute in heritage digitization, where documents include historical handwriting, domain-specific terminology (e.g., jewelry, prehistory, architecture), and non-standard layouts requiring high-dimensional structured extraction.
We present a comparative study of eight open-source lightweight \acp{VLM} ($\leq 7$B parameters) for \ac{OCR}-to-structure across three university heritage collections. Given a document image, models must extract text and generate schema-compliant \ac{JSON}, enabling automatic validation and downstream use. We evaluate models under a constraint-aware protocol across zero-shot, few-shot, and fine-tuning settings, measuring extraction fidelity and structured-output quality using \ac{CER}, \ac{ANLS}$^\ast$, and \ac{mAP-F1}.
To assess practical improvements, we run controlled additional studies on a baseline fine-tuning pipeline, testing the independent impact of (i) hyperparameter optimization, (ii) classical image preprocessing (illumination flattening, denoising, and \ac{CLAHE}), and (iii) multi-stage training, each compared directly to the finetune baseline. Finally, we analyze the trade-off between dataset-specific fine-tuning and a single multi-dataset checkpoint, where joint training enables one model to operate across collections but can shift performance between datasets. Overall, we show that carefully adapted $\leq 7$B \acp{VLM} can provide a sustainable, private, high-performing alternative to manual transcription or commercial black-box systems, and we offer actionable guidance for heritage institutions seeking institution-controlled \ac{OCR}-to-\ac{JSON} extraction.
Code for this paper is available at our project repository.\footnote{\tiny \url{https://github.com/uddipan77/Analysis-of-Lightweight-Vision-Language-Models-for-Document-OCR-and-Structured-Output-Generation}}

\keywords{Document \ac{OCR} \and \acp{VLM} \and structured extraction \and \ac{JSON} generation \and lightweight models \and fine-tuning}
\end{abstract}

\section{Introduction}
\label{sec:intro}

\begin{figure}[tbp]
\centering
\begin{minipage}[t]{0.50\textwidth}
  \vspace{0pt}
  \figplaceholder[\linewidth]{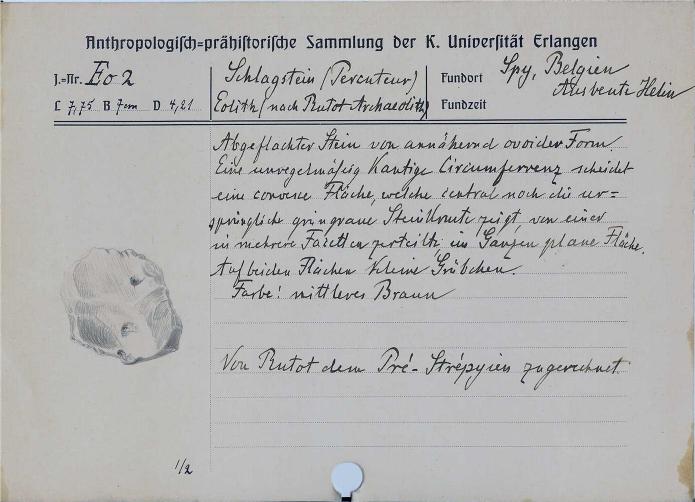}
\end{minipage}\hfill
\begin{minipage}[t]{0.45\textwidth}
  \vspace{0pt}
\IfFileExists{jsons/inventory.json}{%
\begin{lstlisting}[language=json]
  {
    "image_name": string,
    "Überschrift": string,
    "Inventarnummer": string,
    "Maße": {
      "L": string,
      "B": string,
      "D": string
    },
    "Objektbezeichnung": string,
    "Fundort": string,
    "Fundzeit": string,
    "Beschreibungstext": string
  }
\end{lstlisting}%
  }{%
    \lstinputlisting[language=json,basicstyle=\ttfamily\fontsize{8pt}{8pt}\selectfont]{inventory_fallback.json}%
  }
\end{minipage}
\caption{Document example for the Erlangen-Prehistoric dataset (left) and its target \ac{JSON} structure (right). Image from the \enquote{Inventarbuch der Anthropologisch Prähistorischen Sammlung,} courtesy of the Sammlung für Ur- und Frühgeschichte, FAU Erlangen-Nürnberg. Reproduced with permission.}
\label{fig:example}
\end{figure}

\ac{OCR} is no longer only about reading plain text. In many real applications, the goal is to extract information from documents and return it in a structured form that can be searched, validated, or directly used by downstream systems. While standard benchmarks often focus on commercial forms like invoices or ID cards, practitioners in archives, collections, and museums face the distinct challenge of digitizing specialized historical records. In this work, we focus on \ac{OCR}-to-structure: given a document image, the model should output the extracted content as a structured \ac{JSON} object (see \cref{fig:example} for a representative example from an archival inventory).

Recently,\acp{VLM} have shown strong capabilities on document understanding tasks. However, many state-of-the-art models are large, closed-source, and accessible only through commercial cloud \acp{API}. For public archives and collections, this reliance on external providers presents significant barriers regarding data autonomy, i.e., the ability of an institution to maintain sovereignty over its own digital assets. Furthermore, the massive energy consumption of hyperscale models raises concerns about the environmental sustainability of large-scale digitization projects. In practical settings such as on-device systems, cost-sensitive cloud services, or scenarios with limited \ac{GPU} resources, smaller models are often preferred. This motivates our study of lightweight \acp{VLM}: models with up to 7 billion parameters that can be deployed locally on institutional hardware. While these models are attractive for deployment, it is still unclear how reliable they are at producing correct and well-structured \ac{OCR} outputs, especially when the output must follow a highly specialized archival schema.

In this paper, we present a comparative study of eight lightweight \acp{VLM} for structured \ac{OCR}, evaluated on records from three specialized university collections: the Erlangen-Prehistoric dataset (prehistoric artifacts at FAU Erlangen-Nürnberg), the Pforzheim Jewelry dataset (jewelry records at HS Pforzheim), and the Regensburg Scalalogy dataset (architectural staircase forms at OTH Regensburg). We evaluate all models in zero-shot, few-shot, and fine-tuning settings. Importantly, we follow a fair protocol that respects each model's interface: some models only accept a fixed \ac{OCR}-style instruction (e.g., a single keyword prompt such as ``\ac{OCR}''), while others allow custom prompts that explicitly request structured \ac{JSON} extraction. Based on the overall results across zero-shot, few-shot, and fine-tuning evaluations, we select the top three models and investigate further improvements via additional studies. Finally, we analyze single-dataset versus multi-dataset fine-tuning to understand cross-collection transfer and deployment trade-offs.

We investigate practical questions that matter for institutional deployments. First, we study whether multi-stage training provides consistent gains over standard single-stage fine-tuning for \ac{OCR}-to-\ac{JSON} extraction. Second, we test whether hyperparameter optimization improves structured \ac{OCR} performance. Third, we analyze whether classical image preprocessing can help models produce cleaner and more accurate \ac{JSON}.

Overall, our goal is to provide clear guidance for practitioners from \ac{GLAM} who want to use small \acp{VLM} for document \ac{OCR} and structured extraction. We benchmark the models, identify strong lightweight choices, and study practical adaptation strategies that improve reliability without requiring external commercial \acp{API} or excessive resources.

The main contributions of this paper are:
\begin{itemize}
  \item We benchmark eight lightweight \acp{VLM} ($\leq$7B) on three specialized archival collections, evaluating structured \ac{OCR}-to-\ac{JSON} as a private, sustainable alternative to commercial \acp{API}.
  \item We propose a constraint-aware evaluation protocol that accounts for diverse model interfaces (fixed vs.\ custom prompting) in zero-shot settings.
  \item We evaluate the impact of fine-tuning, multi-stage fine-tuning, hyperparameter optimization, and classical image preprocessing on the top-performing models to quantify their effect on extraction accuracy and schema adherence.
\end{itemize}

\section{Related Work}
\label{sec:related}

\textbf{Traditional \ac{OCR} and Document Understanding.}
Classical \ac{OCR} systems such as Tesseract~\cite{tesseract} rely on multi-stage pipelines (e.g., binarization, layout analysis, segmentation, recognition) and are brittle under complex layouts and degraded images. Structured extraction is often addressed by combining \ac{OCR} with downstream \ac{KIE} models. LayoutLM~\cite{layoutlm} and successors LayoutLMv2~\cite{layoutlmv2} and LayoutLMv3~\cite{layoutlmv3} improved multimodal document understanding, but still depend on external \ac{OCR} and typically require task-specific fine-tuning with labeled bounding boxes.

\textbf{End-to-End Document Understanding without \ac{OCR}.}
To remove \ac{OCR} dependence, Donut~\cite{donut} maps document images directly to structured sequences, and Pix2Struct~\cite{pix2struct} pretrains on webpage screenshots to transfer to document tasks. Nougat~\cite{nougat} targets academic documents (PDF-to-Markdown), while DAN~\cite{dan} addresses handwriting recognition without explicit line detection. GOT-\ac{OCR} 2.0~\cite{gotocr} unifies diverse \ac{OCR} tasks and supports structured outputs (e.g., Markdown, \LaTeX). However, these \ac{OCR}-free approaches are typically evaluated primarily for accuracy rather than schema compliance required by institutional databases.

\textbf{Vision-Language Models for Document Tasks.}
General-purpose vision-language models \acp{VLM} have also been applied to document understanding. LLaVA~\cite{llava} popularized visual instruction tuning, and InstructBLIP~\cite{instructblip} and BLIP-2~\cite{blip2} improved vision-language alignment. For documents, mPLUG-DocOwl~\cite{docowl} applied unified instruction tuning, and Qwen-VL~\cite{qwenvl} and successors support high-resolution inputs. Recent models such as Qwen2-VL~\cite{qwen2vl} and InternVL~\cite{internvl} have advanced performance on DocVQA~\cite{docvqa} and FUNSD~\cite{funsd}, but evaluations largely emphasize question answering rather than \ac{JSON} validity and schema compliance. Most recently, the importance of open weights for data autonomy has been increasingly recognized. This has driven the introduction of powerful models like Pixtral~\cite{pixtral} and DeepSeek OCR~\cite{deepseekocr}, which achieve state-of-the-art document understanding and allow for open deployment.

\textbf{Lightweight \acp{VLM} and Efficient Adaptation.}
Resource constraints motivate smaller \acp{VLM} and efficient adaptation. \ac{LoRA}~\cite{lora} and \ac{QLoRA}~\cite{qlora} enable parameter-efficient fine-tuning on accessible hardware. Lightweight \acp{VLM} include PaliGemma~\cite{paligemma}, Phi-3-Vision~\cite{phi35vision}, MiniCPM-V~\cite{minicpm}, SmolVLM~\cite{smolvlm}, and Florence-2~\cite{florence2}. For pure, specialized text extraction, models like PaddleOCR-VL~\cite{paddleocrvl} have recently gained traction. However, their rigid interfaces often do not allow for the complex conversational prompting required to recognize the complex structures inherent in collection records. Furthermore, systematic comparisons of lightweight models under controlled conditions remain scarce, particularly for producing valid structured outputs (e.g., \ac{JSON}) on complex archival records.

\textbf{Research Gap.}
Despite significant progress, several gaps remain in the literature. First, most benchmarks evaluate large-scale models, leaving uncertainty about how lightweight \acp{VLM} ($\leq$7B parameters) perform on structured \ac{OCR} tasks while maintaining institutional data sovereignty. Second, evaluations typically report accuracy metrics without measuring \ac{JSON} validity or schema adherence, which are critical for archival integration. Third, comparisons across models rarely account for interface constraints: some models accept only fixed \ac{OCR} prompts, while others support custom instructions for structured extraction. Fourth, the interaction between classical image preprocessing techniques and modern \acp{VLM} is underexplored. It is unclear whether preprocessing still benefits end-to-end models. Finally, practical adaptation strategies such as multi-stage training and hyperparameter optimization have not been systematically studied for \ac{OCR}-to-\ac{JSON} extraction. Our work addresses these gaps through a controlled benchmark of eight lightweight \acp{VLM}, a fair evaluation protocol respecting model constraints, and studies on preprocessing, hyperparameter optimization, and multi-stage training.

\section{Dataset, Task, and Evaluation}
\label{sec:data_task_eval}
\subsection{Datasets and Splits}
\label{sec:datasets}

We evaluate our pipeline on three distinct datasets sourced from German university collections in archaeology, design history, and architecture. These datasets were developed in close collaboration with the respective institutions to support their ongoing efforts in digitizing historical records. They represent a spectrum of real-world archival challenges, ranging from historical handwritten text to printed catalog cards and complex, nested architectural forms.
\Cref{tab:splits} summarizes the statistics and splits for each collection.

\begin{table}[t]
\caption{Dataset statistics and splits. All datasets are split into 70\,\% training, 15\,\% validation, and 15\,\% test sets.}
\label{tab:splits}
\centering
\footnotesize
\begin{tabular}{lcccccr}
\toprule
\textbf{Dataset} & \textbf{Total} & \textbf{Train} & \textbf{Val} & \textbf{Test} & \textbf{Text Type} & \textbf{Schema Complexity} \\
\midrule
Erlangen-Prehistoric & \phantom{0,}308 & 216 & \phantom{0}46 & \phantom{0}46 & Handwritten & Medium (8 fields) \\
Pforzheim Jewelry & \phantom{0,}590 & 413 & \phantom{0}88 & \phantom{0}89 & Mostly printed & Medium (17 fields) \\
Regensburg Scalalogy & \phantom{0,}164 & 114 & \phantom{0}25 & \phantom{0}25 & Mixed & High (nested, booleans) \\
\midrule
\textbf{Total} & 1{,}062 & 743 & 159 & 160 & --- & --- \\
\bottomrule
\end{tabular}
\end{table}

\textbf{Erlangen-Prehistoric Dataset.}
This dataset originates from the Collection for Prehistory and Protohistory at the Friedrich-Alexander Universität Erlangen-Nürnberg. We processed 308 pages of the historic \enquote{Inventarbuch der Anthropologisch Prähistorischen Sammlung}, which is publicly available through the collection's digital archive.\footnote{\url{https://www.uf.phil.fau.de/abteilungen/sammlung/inventarbuecher/}}
Ground-truth transcriptions were generated using Gemini3 and subsequently manually verified; the transcriptions are published via Zenodo.\footnote{\url{https://zenodo.org/records/20668191}}. 
The documents are predominantly handwritten with printed headers or labels. \Cref{fig:example} shows an example, along with its target data structure. Each record describes archaeological or historical artifacts with fields including inventory number, dimensions, find location, find date, and descriptive text. The handwritten nature of most entries, combined with historical script variations and occasional degradation, makes this dataset particularly challenging for \ac{OCR} systems.

\begin{figure}[tbp]
\centering
\begin{minipage}[t]{0.50\textwidth}
  \vspace{0pt}
  \figplaceholder[\linewidth]{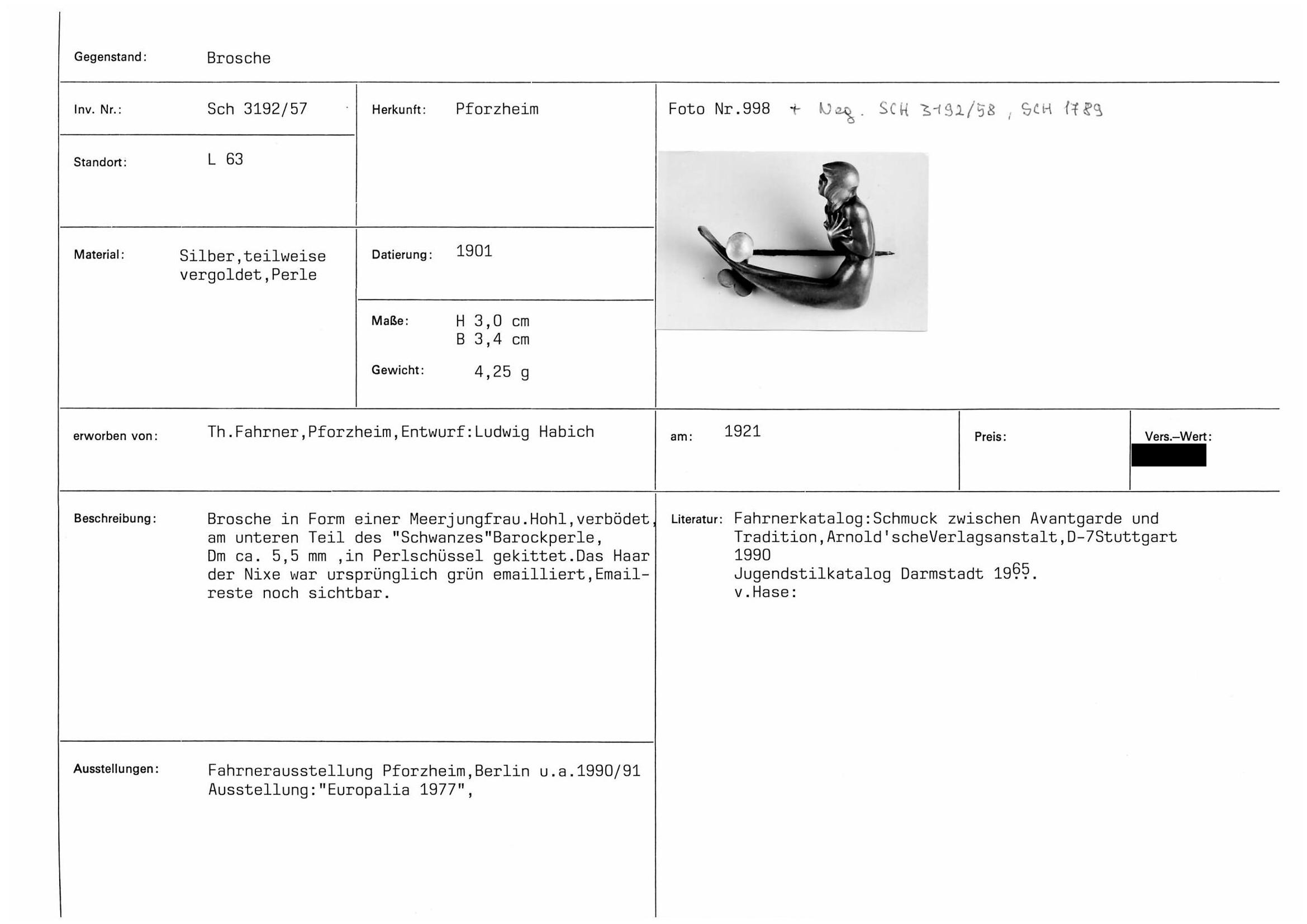}
\end{minipage}\hfill
\begin{minipage}[t]{0.45\textwidth}
  \vspace{0pt}
\IfFileExists{jsons/schmuck.json}{%
\begin{lstlisting}[language=json,basicstyle=\ttfamily\scriptsize]
  { "Gegenstand": string,
    "Inv.Nr": string,
    "Herkunft": string,
    "Foto Notes": string,
    "Standort": string,
    "Material": string,
    "Datierung": string,
    "Maße": string,
    "Gewicht": string,
    "erworben von": string,
    "am": string,
    "Preis": string,
    "Vers.-Wert": string,
    "Beschreibung": string,
    "Literatur": string,
    "Ausstellungen": string,
    "file_name": string  }
\end{lstlisting}%
  }{%
    \lstinputlisting[language=json,basicstyle=\ttfamily\scriptsize]{schmuck_fallback.json}%
  }
\end{minipage}
\caption{Representative document example for the jewelry dataset (left) and its target \ac{JSON} structure (right). Image courtesy of Pforzheim Jewellery Museum and HS Pforzheim (KUPFER Projekt). Reproduced with permission.}
\label{fig:jewelry}
\end{figure}

\textbf{Pforzheim Jewelry Dataset.}
This dataset comprises 590 digitized inventory cards from the historical teaching aids collection at the Pforzheim University, School of Design. The records come from the KUPFER research project, which investigates the collection's history and significance.\footnote{KUPFER--Die Lehrmittelsammlung der Kunstgewerbeschule Pforzheim zwischen Medienarchäologie und Innovationstransfer (2023--2027), funded by the Federal Ministry of Research, Technology and Space (BMFTR) (\url{https://designpf.hs-pforzheim.de/kupfer})}
They describe jewelry from different periods and cultural areas, acquired by the Pforzheim School of Decorative Arts between 1879 and 1970. The inventory cards were derived from historical books when the objects were moved on permanent loan to the Pforzheim Jewellery Museum around 1980. The task requires extracting a dense 17-field schema, including material, weight, and insurance value. An example is provided in \cref{fig:jewelry}.

\begin{figure}[tbp]
\centering
\begin{minipage}[t]{0.40\textwidth}
  \vspace{0pt}
  \figplaceholder[\linewidth]{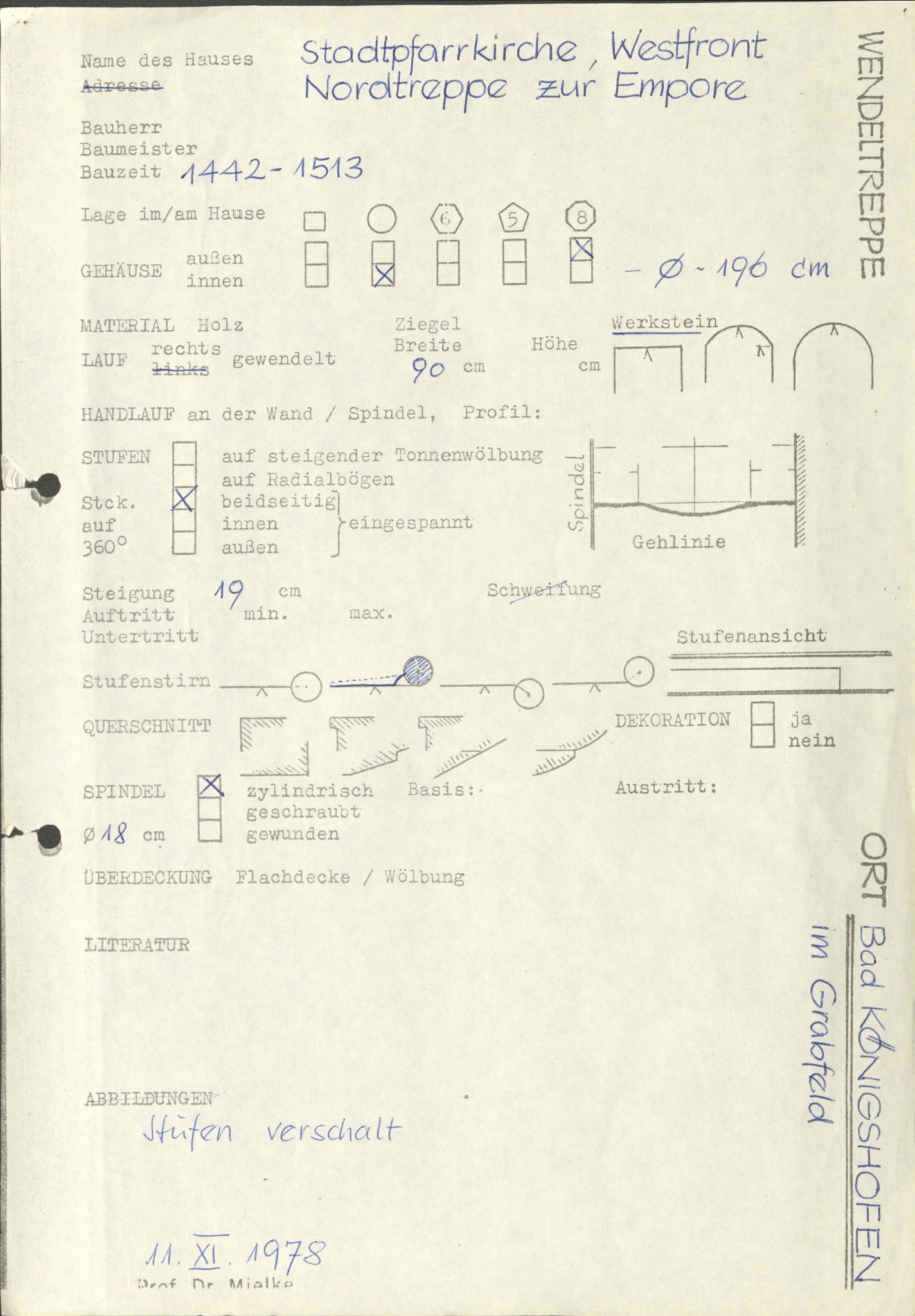}
\end{minipage}\hfill
\begin{minipage}[t]{0.58\textwidth}
  \vspace{0pt}
\IfFileExists{jsons/staircase_type1.json}{%
\begin{lstlisting}[language=json,basicstyle=\ttfamily\tiny]
  { "stair_type": string, "Name des Hauses": string, "Adresse": string, "Bauherr": string, "Baumeister": string, "Bauzeit": string,
    "LAGE": {"am Hause": bool,"im Hause": bool},
    "Gesamt Durchmesser cm": int,
    "GEHÄUSE": {
      "außen": {"1": bool, "2": bool, "3": bool, "4": bool, "5": bool, "6": bool}, 
      "innen": {"1": bool, "2": bool, "3": bool, "4": bool, "5": bool, "6": bool}},
    "MATERIAL": { "Holz": bool,"Ziegel": bool,"Werkstein": bool },
    "LAUF": {"rechts drehend": bool,"links drehend": bool},
    "Breite cm": integer, "Höhe cm": integer,
    "Ceiling Shape": {"1": bool, "2": bool, "3": bool, "4": bool},
    "HANDLAUF": {"an der Wand": bool, "an der Spindel": bool, "Profil": string},
    "STUFEN": {"Stück auf 360": integer, "Stück insgesamt": integer, "auf steigender Tonnenwölbung": bool, "auf Radialbögen": bool, "beidseitig": bool, "innen": bool, "außen": bool},
    "Notes": string, "Steigung cm": integer, "Schweifung": string, "cm Stich": integer,
    "Auftritt": {"cm min": integer, "cm max": integer},
    "Untertritt cm": integer,
    "Distanzkerbe": {"ja": bool, "nein": bool},
    "Stufenstirn": {"1": bool, "2": bool, "3": bool, "4": bool},
    "Querschnitt obere": {"1": bool, "2": bool, "3": bool},
    "Querschnitt untere": {"1": bool, "2": bool, "3": bool, "4": bool, "5": bool},
    "SPINDEL": {"zylindrisch": bool, "geschraubt": bool, "gewunden": bool, "Durchmesser des Auges cm": integer, "Durchmesser cm": integer},
    "Kapitell": string, "Basis": string,
    "ÜBERDECKUNG des Gehäuses": {"Flachdecke": bool, "Wölbung": bool},
    "DEKORATION": {"ja": bool, "nein": bool},
    "Ort":string,"Datum":string,"Approved by":string,"image_name":string }
\end{lstlisting}%
  }{%
    \lstinputlisting[language=json,basicstyle=\ttfamily\tiny]{staircase_fallback.json}%
  }
\end{minipage}
\caption{Representative document example for the Staircase dataset (left) and its extensive target \ac{JSON} structure (right). Image courtesy of the Friedrich-Mielke-Institut für Scalalogie, OTH Regensburg, Dossier \enquote{Stadtpfarrkirche Bad Königshofen im Grabfeld}. Reproduced with permission.}
\label{fig:staircase_example}
\end{figure}

\textbf{Regensburg Scalalogy Dataset.}
This dataset consists of 164 survey forms from the Scalalogy Collections of the Friedrich Mielke Institute (FMIS) at OTH Regensburg.\footnote{\url{https://architektur.oth-regensburg.de/forschung/scalalogie}}
See \cref{fig:staircase_example} for an example. These records form the methodological backbone of "Scalalogy", the scientific study of stairs, and were created by Friedrich Mielke to systematically document stairs using a standardized nomenclature. The forms contain a mixture of printed field labels and handwritten entries for various staircase types (e.g., spiral and interior). With a target \ac{JSON} structure featuring deeply nested objects and numerous boolean fields for technical attributes, this dataset serves as a high-complexity stress test for hierarchical schema compliance.

\subsection{Task Definition and Metrics}
\label{sec:task_metrics}

\textbf{Task Definition.}
Given a document image, the model must output a \ac{JSON} object that follows a fixed, dataset-specific schema. Erlangen-Prehistoric dataset and Pforzheim Jewelry dataset use flat key-value dictionaries with string values, while Regensburg Scalalogy dataset uses a larger schema with nested objects and boolean fields. If a field is missing or illegible, we treat it as empty: text fields should be an empty string and boolean fields should be false. The \ac{JSON} output can be deterministically rendered to Markdown for inspection and manual auditing.

\textbf{Evaluation metrics.}
We report \ac{CER}, \ac{ANLS}$^\ast$, and \ac{mAP-F1}, and additionally track \ac{JSON} validity (parsable dict outputs). Collectively, these metrics disentangle content quality from structured output reliability: \ac{CER} and \ac{ANLS}$^\ast$ measure how well the extracted values match the reference (even when formatting degrades), whereas \ac{mAP-F1} and \ac{JSON} validity quantify whether the output is a usable \ac{JSON} object under a strict parsing requirement.

\textbf{Canonical ground truth and parsing policy.}
For each sample, the ground truth is stored as a fixed canonical \ac{JSON} string (consistent formatting and a fixed key order). For a model output, we first attempt to parse the raw generation as \ac{JSON}. If parsing succeeds, metrics are computed on the parsed object, and for string-based metrics we serialize the parsed prediction into the same canonical \ac{JSON} format as the ground truth. If parsing fails, we compare the raw prediction text directly against the canonical ground-truth \ac{JSON} string, reflecting downstream usability requirements.

\textbf{\ac{CER}.}
We compute \ac{CER} against a fixed canonical ground-truth \ac{JSON} string (with a consistent key order and formatting as stored). If the model output parses as \ac{JSON}, we serialize the parsed prediction into the same canonical format and compute \ac{CER} between the two \ac{JSON} strings. If parsing fails, we compute \ac{CER} between the raw model output text and the canonical ground-truth \ac{JSON} string. Lower \ac{CER} indicates higher overall output fidelity.

\textbf{\ac{ANLS}$^\ast$.}
If both prediction and ground truth parse as \ac{JSON}, we apply structured \ac{ANLS}$^\ast$~\cite{anlsstar} directly to the nested objects. These are then scored with \ac{ANLS} similarity using threshold $\tau = 0.5$ (scores below $\tau$ are set to 0). If the prediction is not valid \ac{JSON}, we fall back to a string-level \ac{ANLS}-style similarity between the raw output text and the canonical ground-truth \ac{JSON} string (fixed order/format), penalizing invalid structured outputs while still giving a graded content score.

\textbf{\ac{mAP-F1}.}
Inspired by COCO-style threshold averaging~\cite{coco} and DAN's~\cite{dan} mAPCER, \ac{mAP-F1} is our structure-first metric. When the prediction is valid \ac{JSON}, we flatten prediction and ground truth into dot-notated leaf keys and compute per-field overlap $1 - \mathrm{CER}(gt, pred)$ at thresholds $\theta \in \{0.3, 0.5, 0.7, 0.8\}$. We choose multiple thresholds to reflect different tolerance levels to \ac{OCR} noise, so the score rewards both approximate and near-exact field matches rather than depending on a single cutoff. Empty GT/non-empty predictions count as FP; for non-empty GT, overlap $\geq \theta$ is TP, otherwise FN (and additionally FP if the prediction is non-empty). We compute precision and recall per field, macro-average them across fields to obtain $P(\theta)$ and $R(\theta)$, set $F1(\theta) = \frac{2P(\theta)R(\theta)}{P(\theta)+R(\theta)}$, and report \ac{mAP-F1} as the mean of $F1(\theta)$ over the four thresholds. If the output is not valid \ac{JSON}, all non-empty GT fields are counted as FN, yielding \ac{mAP-F1} = 0.

\textbf{\ac{JSON} validity and parsing policy.}
We additionally track the \ac{JSON}-validity rate, i.e., the fraction of outputs that parse (via \texttt{json.loads}) into a Python dict. For space constraints, we do not include \ac{JSON}-validity columns in the main result tables. Across our experiments, \ac{JSON} validity stays in the range of $\approx$ 95\,\%--100\,\% for Gemma-3, Phi-3.5-Vision, and Qwen2.5-VL across settings, but remains much lower for Donut, Florence-2, and PaddleOCR-VL even after fine-tuning. All metrics share the same initial parsing step, however, only \ac{CER} and \ac{ANLS}$^\ast$ fall back to raw-text comparison when \ac{JSON} parsing fails, while \ac{mAP-F1} requires valid \ac{JSON} and is set to 0 otherwise.

\section{Experimental Setup and Methods}
\label{sec:methods}

\subsection{Models}
\label{sec:models}

We evaluate eight vision-language models (200M--7B parameters) spanning encoder-decoder and instruction-tuned architectures (\Cref{tab:models}). Models marked with $\dagger$ only accept fixed \ac{OCR}-style prompts and therefore do not support few-shot demonstrations.

\begin{table}[t]
\caption{Overview of the eight VLMs evaluated. Models marked with $\dagger$ only accept fixed OCR-style prompts and do not support few-shot demonstrations.}
\label{tab:models}
\centering
\footnotesize
\begin{tabular}{llr}
\toprule
\textbf{Model} & \textbf{Architecture Type} & \textbf{Parameters} \\
\midrule
Donut-Base\textsuperscript{$\dagger$}~\cite{donut} & Encoder-decoder & 200M \\
Donut-CORD\textsuperscript{$\dagger$}~\cite{donut} & Encoder-decoder & 200M \\
Florence-2-Large\textsuperscript{$\dagger$}~\cite{florence2} & Unified vision-language & 0.77B \\
Nanonets-OCR-s~\cite{nanonets} & OCR-specialized VLMs & 4B \\
PaddleOCR-VL\textsuperscript{$\dagger$}~\cite{paddleocrvl} & OCR-specialized VLMs & 0.9B \\
Gemma-3-4B~\cite{gemma3} & Instruction-tuned VLMs & 4B \\
Phi-3.5-Vision~\cite{phi35vision} & Instruction-tuned VLMs & 4.2B \\
Qwen2.5-VL-7B~\cite{qwen25vl} & Instruction-tuned VLMs & 7B \\
\bottomrule
\end{tabular}
\end{table}

\subsection{Constraint-aware prompting}
\label{sec:prompting}

A key challenge in benchmarking diverse \acp{VLM} is that they differ in their input interfaces and prompting flexibility. Donut-Base, Donut-CORD, Florence-2-Large, and PaddleOCR-VL are restricted to fixed \ac{OCR}-style prompts or task-specific tokens (e.g., \texttt{<s\_cord-v2>}, \texttt{<OCR>}) and do not support arbitrary natural-language instructions.

In contrast, Nanonets-OCR-s, Gemma-3-4B, Phi-3.5-Vision, and Qwen2.5-VL-7B accept custom prompts, enabling schema-guided structured extraction. To ensure a fair comparison, we evaluate each model under its supported interface. For fixed-prompt models, we apply minimal post-processing to raw generations (strip wrapper text/chat prefixes, trim whitespace and code fences) before attempting \ac{JSON} parsing.

\subsection{Zero-shot inference}
\label{sec:zero_shot}

In the zero-shot setting, we provide no in-context demonstrations. For prompt-flexible models, the input prompt includes the target \ac{JSON} skeleton (i.e., the exact field structure used in the ground truth) and requests that the output follows this schema. For fixed-prompt models, we use their designated \ac{OCR} prompt tokens/keywords and evaluate the resulting generations after the same minimal post-processing. All evaluations use greedy decoding (temperature = 0) for deterministic outputs, with images resized according to each model's native preprocessing.

\subsection{Few-shot inference}
\label{sec:few_shot}

Few-shot evaluation is applied only to models that support custom prompting (Nanonets-OCR-s, Qwen2.5-VL, Phi-3.5-Vision, and Gemma-3). We prepend $k \in \{1,2\}$ image-\ac{JSON} demonstrations sampled from the training set:
\begin{lstlisting}[basicstyle=\ttfamily\scriptsize]
[Image 1] -> {"field1": "value1", ...}
[Image 2] -> {"field1": "value1", ...}
[Query Image] ->
\end{lstlisting}

\noindent In our experiments, $k=1$ is used for the Pforzheim Jewelry dataset, while $k=2$ is used for the Erlangen-Prehistoric and Regensburg Scalalogy datasets.

\subsection{Standard fine-tuning}
\label{sec:fine_tuning}

We fine-tune all evaluated models. For models below 1B parameters (Donut-Base, Donut-CORD, Florence-2-Large, and PaddleOCR-VL), we perform full-parameter fine-tuning. For larger models (Nanonets-OCR-s, Gemma-3-4B, Phi-3.5-Vision, and Qwen2.5-VL-7B), we use parameter-efficient \ac{QLoRA}~\cite{qlora} on an NVIDIA V100 (32GB \ac{VRAM}), combining 4-bit quantization with \ac{LoRA} adapters while keeping the base weights frozen. \ac{LoRA} adapters are added to the attention and MLP layers. The best checkpoint is selected by lowest autoregressive validation \ac{CER} and evaluated on the test set via autoregressive generation.

\subsection{Additional Study Configurations}
\label{sec:additional_studies}

Using the fine-tuning pipeline described above as the baseline, we conduct three additional studies to evaluate practical strategies for improving structured \ac{OCR} performance. These studies are applied only to the selected top-3 models (selection criteria reported in \Cref{sec:results}) and are evaluated independently to isolate each contribution.

\textbf{Study 1: Hyperparameter Optimization.}
We compare the default fine-tuning configuration against an Optuna~\cite{optuna}-optimized configuration for each of the top-3 models. Optuna is used only to identify the best hyperparameters; these values are then plugged back into the same baseline fine-tuning code for the final training run. We run 20 trials using the \ac{TPE} sampler, optimizing validation \ac{CER}. To reduce computational cost, each trial trains for 5 epochs, after which the best hyperparameters are used for a full training run.

\textbf{Study 2: Image Preprocessing.}
We evaluate whether classical image preprocessing improves model performance compared to using raw images. Our preprocessing pipeline consists of four sequential steps: 1. Illumination Flattening: Convert to grayscale, estimate background via morphological opening, and normalize to remove uneven lighting. 2. Denoising: Apply bilateral filtering (default) or Non-Local Means denoising to reduce noise while preserving edges. 3. \ac{CLAHE}: Apply \ac{CLAHE} on the luminance channel to enhance local contrast. 4. Letterbox Resize: Resize the image to a fixed canvas (e.g., $1024 \times 1024$) with padding to preserve aspect ratio. We apply the full preprocessing pipeline on top of the baseline fine-tuning setup and compare performance against the same training/evaluation run using raw images.

\textbf{Study 3: Multi-stage Training.}
We investigate whether curriculum-based multi-stage training~\cite{curriculum} improves convergence and final performance compared to single-stage fine-tuning. The procedure is: 1. Warm-up: Fine-tune on the target dataset's training split for a short warm-up phase using a higher learning rate (i.e., $3\times$ the base LR). This stage uses teacher forcing (supervised \ac{JSON} targets) to quickly adapt the model to the document domain and the dataset-specific \ac{JSON} style. 2. Refinement: Continue fine-tuning on the same dataset with the base learning rate for the remaining epochs. We run validation at each epoch and select the best model by the lowest validation \ac{CER} (early stopping applied when validation \ac{CER} stops improving for a patience period of $N$ epochs). We compare this two-stage schedule against a single-stage baseline trained for the same total number of epochs using cosine learning-rate decay.

\section{Results}
\label{sec:results}

\subsection{Benchmark of eight VLMs Across Settings}
\label{sec:benchmark_results}

Tables~\ref{tab:zero_shot} to~\ref{tab:fine_tuning_results} summarize the performance of the evaluated models on the three datasets (Erlangen-Prehistoric, Pforzheim Jewelry, Regensburg Scalalogy) under zero-shot, few-shot (1--2 examples), and fine-tuning settings, respectively. We report \ac{CER}, \ac{ANLS}$^\ast$, and \ac{mAP-F1}. Consequently, a model can achieve low \ac{CER}/high \ac{ANLS}$^\ast$ yet still obtain a lower \ac{mAP-F1} if values are assigned to the wrong keys, fields are missing/merged, or extra fields are hallucinated. Notably, Donut can achieve lower \ac{CER}/\ac{ANLS}$^\ast$ after fine-tuning, yet still yields 0 \ac{mAP-F1} because its outputs often fail strict \ac{JSON} dict parsing (schema-incompatible/malformed).

\begin{table}[tbp]
\caption{Zero-shot results for all eight models across three datasets. All metrics are reported as percentages. $\downarrow$: lower is better; $\uparrow$: higher is better.}
\label{tab:zero_shot}
\centering
\scriptsize
\begin{tabular}{lccccccccr}
\toprule
\multirow{2}{*}{\textbf{Model}} & \multicolumn{3}{c}{\textbf{Erlangen-Prehistoric}} & \multicolumn{3}{c}{\textbf{Pforzheim Jewelry}} & \multicolumn{3}{c}{\textbf{Regensburg Scalalogy}} \\
\cmidrule(lr){2-4}\cmidrule(lr){5-7}\cmidrule(lr){8-10}
& CER$\downarrow$ & ANLS*$\uparrow$ & mAP-F1$\uparrow$ & CER$\downarrow$ & ANLS*$\uparrow$ & mAP-F1$\uparrow$ & CER$\downarrow$ & ANLS*$\uparrow$ & mAP-F1$\uparrow$ \\
\midrule
Donut-Base$^\dagger$ & 93.4 & \phantom{0}0.0 & \phantom{0}0.0 & 97.2 & \phantom{0}0.0 & \phantom{0}0.0 & 90.1 & \phantom{0}0.0 & \phantom{0}0.0 \\
Donut-CORD$^\dagger$ & 78.8 & \phantom{0}0.0 & \phantom{0}0.0 & 91.9 & \phantom{0}0.0 & \phantom{0}0.0 & 93.1 & \phantom{0}0.0 & \phantom{0}0.0 \\
Florence-2-Large$^\dagger$ & 45.6 & 50.4 & \phantom{0}0.0 & 33.6 & 66.4 & \phantom{0}0.0 & 75.6 & \phantom{0}0.0 & \phantom{0}0.0 \\
Nanonets-OCR-s & 25.7 & 49.4 & 28.5 & \phantom{0}1.9 & 86.5 & 86.5 & 39.3 & 22.8 & 15.2 \\
PaddleOCR-VL$^\dagger$ & 69.2 & 20.1 & \phantom{0}0.0 & 51.3 & 57.0 & \phantom{0}0.0 & 78.1 & \phantom{0}0.0 & \phantom{0}0.0 \\
Gemma-3-4B & 33.6 & 41.8 & 34.6 & \phantom{0}4.4 & 92.1 & 87.4 & 37.6 & 26.3 & 22.5 \\
Phi-3.5-Vision & 50.9 & 28.1 & 23.4 & \phantom{0}7.3 & 85.3 & 67.9 & 28.7 & 57.1 & 27.9 \\
Qwen2.5-VL-7B & \textbf{21.4} & \textbf{54.0} & \textbf{56.2} & \textbf{\phantom{0}2.6} & \textbf{96.0} & \textbf{96.3} & \textbf{21.3} & \textbf{67.0} & \textbf{33.3} \\
\bottomrule
\end{tabular}
\end{table}

\begin{table}[tbp]
\caption{Few-shot results (1--2 examples) for models that support custom prompting and in-context demonstrations. Models restricted to fixed OCR prompts (Donut-Base, Donut-CORD, Florence-2-Large, PaddleOCR-VL) are excluded. All metrics are reported as percentages.}
\label{tab:few_shot}
\centering
\scriptsize
\begin{tabular}{lccccccccr}
\toprule
\multirow{2}{*}{\textbf{Model}} & \multicolumn{3}{c}{\textbf{Erlangen-Prehistoric}} & \multicolumn{3}{c}{\textbf{Pforzheim Jewelry}} & \multicolumn{3}{c}{\textbf{Regensburg Scalalogy}} \\
\cmidrule(lr){2-4}\cmidrule(lr){5-7}\cmidrule(lr){8-10}
& CER$\downarrow$ & ANLS*$\uparrow$ & mAP-F1$\uparrow$ & CER$\downarrow$ & ANLS*$\uparrow$ & mAP-F1$\uparrow$ & CER$\downarrow$ & ANLS*$\uparrow$ & mAP-F1$\uparrow$ \\
\midrule
Nanonets-OCR-s & 61.9 & 53.3 & 44.7 & \phantom{0}1.7 & 93.1 & 90.7 & 33.3 & 44.5 & \textbf{44.9} \\
Gemma-3-4B & 26.8 & 61.7 & 53.6 & \phantom{0}2.0 & 97.3 & 92.1 & 34.9 & 33.8 & 20.2 \\
Phi-3.5-Vision & 59.3 & 40.7 & 38.9 & \phantom{0}7.2 & 85.9 & 71.1 & 52.7 & 27.5 & 23.1 \\
Qwen2.5-VL-7B & \textbf{18.9} & \textbf{79.0} & \textbf{74.7} & \textbf{\phantom{0}0.9} & \textbf{98.9} & \textbf{99.4} & \textbf{24.1} & \textbf{55.6} & 30.7 \\
\bottomrule
\end{tabular}
\end{table}

\begin{table}[tbp]
\caption{Fine-tuning results for all eight models across three datasets. All metrics are reported as percentages.}
\label{tab:fine_tuning_results}
\centering
\scriptsize
\begin{tabular}{lccccccccr}
\toprule
\multirow{2}{*}{\textbf{Model}} & \multicolumn{3}{c}{\textbf{Erlangen-Prehistoric}} & \multicolumn{3}{c}{\textbf{Pforzheim Jewelry}} & \multicolumn{3}{c}{\textbf{Regensburg Scalalogy}} \\
\cmidrule(lr){2-4}\cmidrule(lr){5-7}\cmidrule(lr){8-10}
& CER$\downarrow$ & ANLS*$\uparrow$ & mAP-F1$\uparrow$ & CER$\downarrow$ & ANLS*$\uparrow$ & mAP-F1$\uparrow$ & CER$\downarrow$ & ANLS*$\uparrow$ & mAP-F1$\uparrow$ \\
\midrule
Donut-Base & 16.4 & 78.0 & \phantom{0}0.0 & \phantom{0}6.3 & 93.6 & \phantom{0}0.0 & 59.5 & \phantom{0}0.0 & \phantom{0}0.0 \\
Donut-CORD & 20.6 & 75.0 & \phantom{0}0.0 & \phantom{0}6.9 & 92.9 & \phantom{0}0.0 & 66.0 & \phantom{0}0.0 & \phantom{0}0.0 \\
Florence-2-Large & 10.7 & 75.5 & 45.4 & 11.1 & 89.9 & \phantom{0}0.0 & 57.4 & \phantom{0}4.3 & \phantom{0}0.0 \\
Nanonets-OCR-s & 23.0 & 82.5 & 81.2 & \phantom{0}0.5 & 96.7 & 87.2 & 20.2 & 79.9 & 50.7 \\
PaddleOCR-VL & 13.7 & 90.1 & 85.3 & \phantom{0}1.1 & 98.7 & 84.7 & 65.1 & \phantom{0}0.0 & \phantom{0}0.0 \\
Gemma-3-4B & 10.4 & 88.6 & 83.1 & \phantom{0}0.9 & 99.0 & 99.4 & 14.6 & 73.8 & 46.9 \\
Phi-3.5-Vision & 17.9 & 86.5 & 82.2 & \textbf{\phantom{0}0.3} & \textbf{99.4} & \textbf{99.5} & \textbf{\phantom{0}6.3} & \textbf{90.1} & \textbf{72.0} \\
Qwen2.5-VL-7B & \textbf{\phantom{0}5.8} & \textbf{91.0} & \textbf{90.0} & \phantom{0}0.5 & 99.2 & 99.4 & \phantom{0}7.6 & 87.0 & 62.8 \\
\bottomrule
\end{tabular}
\end{table}

\textbf{Top-3 selection.}
Based on the overall results across the three settings (Tables~\ref{tab:zero_shot}--\ref{tab:fine_tuning_results}), we select the top three models, Phi-3.5-Vision, Qwen2.5-VL-7B, and Gemma-3-4B, as they show the strongest potential for reliable structured extraction (low \ac{CER}, high \ac{ANLS}$^\ast$, and high \ac{mAP-F1}) and stable \ac{JSON} generation across datasets. Across zero-shot, few-shot, and fine-tuning, Phi-3.5-Vision, Qwen2.5-VL-7B, and Gemma-3-4B maintain near-saturated \ac{JSON}-validity ($\approx$ 95\,\%--100\,\%) across datasets, reflecting stable structured generation. In the next subsection, we report additional studies and practical improvements on these selected models.

\subsection{Secondary Studies on the Selected Top-3 Models}
\label{sec:secondary_studies}

We run three additional experiments on the selected top-3 models to test practical strategies for improving structured \ac{OCR} performance. Each study is applied independently to isolate its contribution.

\textbf{Study 1: Hyperparameter Optimization.}
\Cref{tab:hpo} compares default fine-tuning settings against the best settings found using Optuna for each of the top-3 models.

\begin{table}[tbp]
\caption{Effect of hyperparameter optimization on the selected top-3 models across three datasets. Scores are reported after fine-tuning. All metrics are reported as percentages.}
\label{tab:hpo}
\centering
\smalltablefont
\begin{tabular}{lccccccccr}
\toprule
\multirow{2}{*}{\textbf{Model / Setting}} & \multicolumn{3}{c}{\textbf{Erlangen-Prehistoric}} & \multicolumn{3}{c}{\textbf{Pforzheim Jewelry}} & \multicolumn{3}{c}{\textbf{Regensburg Scalalogy}} \\
\cmidrule(lr){2-4}\cmidrule(lr){5-7}\cmidrule(lr){8-10}
& CER$\downarrow$ & ANLS*$\uparrow$ & mAP-F1$\uparrow$ & CER$\downarrow$ & ANLS*$\uparrow$ & mAP-F1$\uparrow$ & CER$\downarrow$ & ANLS*$\uparrow$ & mAP-F1$\uparrow$ \\
\midrule
Gemma-3-4B (Default) & 10.4 & 88.6 & 83.1 & 0.9 & 99.0 & 99.4 & 14.6 & 73.8 & 46.9 \\
Gemma-3-4B (HPO) & \phantom{0}9.3 & 88.8 & 65.4 & 0.4 & 99.2 & 99.5 & \phantom{0}3.4 & 92.3 & 77.9 \\
Phi-3.5-Vision (Default) & 17.9 & 86.5 & 82.2 & 0.3 & 99.4 & 99.5 & \phantom{0}6.3 & 90.1 & 72.0 \\
Phi-3.5-Vision (HPO) & 17.7 & 84.0 & 81.2 & 0.3 & 99.5 & 99.7 & \phantom{0}3.2 & 91.8 & 73.8 \\
Qwen2.5-VL-7B (Default) & \phantom{0}5.8 & 91.0 & 90.0 & 0.5 & 99.2 & 99.4 & \phantom{0}7.6 & 87.0 & 62.8 \\
Qwen2.5-VL-7B (HPO) & \phantom{0}4.3 & 93.4 & 91.2 & 0.5 & 99.3 & 99.5 & \phantom{0}3.1 & 89.4 & 67.1 \\
\bottomrule
\end{tabular}
\end{table}

Hyperparameter optimization generally improves performance, but the effect is not uniform across models, datasets, or metrics. The largest gains appear on the Regensburg Scalalogy dataset for all three models, while improvements on Pforzheim Jewelry dataset are comparatively small as they were already quite accurate. On Erlangen-Prehistoric dataset, hyperparameter optimization clearly benefits Qwen2.5-VL-7B, but the effect is mixed for Gemma-3-4B and Phi-3.5-Vision (e.g., some metrics improve while \ac{mAP-F1} degrade). Overall, hyperparameter optimization is a strong enhancement in several settings, but it should be evaluated empirically rather than assumed to provide consistent gains.

\textbf{Study 2: Image Preprocessing.}
\Cref{tab:preprocessing} reports the effect of classical preprocessing compared to raw images for the selected top-3 models.

\begin{table}[tbp]
\caption{Effect of image preprocessing on the selected top-3 models across three datasets. Scores are reported after fine-tuning. All metrics are reported as percentages.}
\label{tab:preprocessing}
\centering
\smalltablefont
\begin{tabular}{lcccccccccc}
\toprule
\multirow{2}{*}{\textbf{Model}} & \multirow{2}{*}{\textbf{Preproc.}} & \multicolumn{3}{c}{\textbf{Erlangen-Prehistoric}} & \multicolumn{3}{c}{\textbf{Pforzheim Jewelry}} & \multicolumn{3}{c}{\textbf{Regensburg Scalalogy}} \\
\cmidrule(lr){3-5}\cmidrule(lr){6-8}\cmidrule(lr){9-11}
& & CER$\downarrow$ & ANLS*$\uparrow$ & mAP-F1$\uparrow$ & CER$\downarrow$ & ANLS*$\uparrow$ & mAP-F1$\uparrow$ & CER$\downarrow$ & ANLS*$\uparrow$ & mAP-F1$\uparrow$ \\
\midrule
Gemma-3-4B & & 10.4 & 88.6 & 83.1 & 0.9 & 99.0 & 99.4 & 14.6 & 73.8 & 46.9 \\
Gemma-3-4B & $\checkmark$ & 10.4 & 88.8 & 85.3 & 1.2 & 97.6 & 97.7 & 16.4 & 69.7 & 43.6 \\
Phi-3.5-Vision & & 17.9 & 86.5 & 82.2 & 0.3 & 99.4 & 99.5 & \phantom{0}6.3 & 90.1 & 72.0 \\
Phi-3.5-Vision & $\checkmark$ & 19.0
 & 80.4 & 76.5 & 0.3 & 99.2 & 99.3 & \phantom{0}5.5 & 89.5 & 61.1 \\
Qwen2.5-VL-7B & & \phantom{0}5.8 & 91.0 & 90.0 & 0.5 & 99.2 & 99.4 & \phantom{0}7.6 & 87.0 & 62.8 \\
Qwen2.5-VL-7B & $\checkmark$ & \phantom{0}5.2 & 90.0 & 87.4 & 0.7 & 99.1 & 99.3 & 12.0 & 81.7 & 53.6 \\
\bottomrule
\end{tabular}
\end{table}

Overall, image preprocessing shows a mixed, model- and dataset-dependent effect (Table~\ref{tab:preprocessing}). It improves \ac{CER} for Phi-3.5-Vision on the Regensburg Scalalogy dataset (6.3\,\% $\rightarrow$ 5.5\,\%) and for Qwen2.5-VL-7B on the Erlangen-Prehistoric dataset (5.8\,\% $\rightarrow$ 5.2\,\%), but these gains do not consistently translate into better structured extraction (e.g., Phi-3.5-Vision on Regensburg Scalalogy shows lower \ac{mAP-F1} after preprocessing). In several other settings preprocessing is clearly harmful---most notably on the Erlangen-Prehistoric dataset for Phi-3.5-Vision, the Pforzheim Jewelry dataset for Gemma-3-4B and Qwen2.5-VL-7B, and the Regensburg Scalalogy dataset for Gemma-3-4B and Qwen2.5-VL-7B.

\textbf{Study 3: Multi-stage Training.}
\Cref{tab:multistage} compares single-stage fine-tuning against two-stage curriculum training for the selected top-3 models.

\begin{table}[tbp]
\caption{Effect of multi-stage training on the selected top-3 models across three datasets. All metrics are reported as percentages.}
\label{tab:multistage}
\centering
\smalltablefont
\begin{tabular}{lcccccccccr}
\toprule
\multirow{2}{*}{\textbf{Model}} & \multirow{2}{*}{\textbf{Stages}} & \multicolumn{3}{c}{\textbf{Erlangen-Prehistoric}} & \multicolumn{3}{c}{\textbf{Pforzheim Jewelry}} & \multicolumn{3}{c}{\textbf{Regensburg Scalalogy}} \\
\cmidrule(lr){3-5}\cmidrule(lr){6-8}\cmidrule(lr){9-11}
& & CER$\downarrow$ & ANLS*$\uparrow$ & mAP-F1$\uparrow$ & CER$\downarrow$ & ANLS*$\uparrow$ & mAP-F1$\uparrow$ & CER$\downarrow$ & ANLS*$\uparrow$ & mAP-F1$\uparrow$ \\
\midrule
Gemma-3-4B  & single  & 10.4 & 88.6 & 83.1 & 0.9 & 99.0 & 99.4 & 14.6 & 73.8 & 46.9 \\
Gemma-3-4B & multi & \phantom{0}8.1 & 89.9 & 86.5 & 0.4 & 99.3 & 99.5 & 10.3 & 80.2 & 63.8 \\
Phi-3.5-Vision & single & 17.9 & 86.5 & 82.2 & 0.3 & 99.4 & 99.5 & \phantom{0}6.3 & 90.1 & 72.0 \\
Phi-3.5-Vision & multi & 18.0 & 86.7 & 83.0 & 0.3 & 99.5 & 99.7 & \phantom{0}4.9 & 90.5 & 72.5 \\
Qwen2.5-VL-7B & single & \phantom{0}5.8 & 91.0 & 90.0 & 0.5 & 99.2 & 99.4 & \phantom{0}7.6 & 87.0 & 62.8 \\
Qwen2.5-VL-7B & multi & \phantom{0}5.8 & 91.4 & 89.1 & 0.7 & 88.5 & 90.2 & \phantom{0}7.8 & 86.0 & 57.8 \\
\bottomrule
\end{tabular}
\end{table}
In our experiments, multi-stage fine-tuning does not inherently guarantee improved accuracy; its impact is model- and dataset-dependent. From Table~\ref{tab:multistage}, we observe that it consistently benefits Gemma-3-4B across all three datasets. For Phi-3.5-Vision, multi-stage notably improves Regensburg Scalalogy performance, while changes on Erlangen-Prehistoric and Pforzheim Jewelry are small and can be negative depending on the metric. For Qwen2.5-VL-7B, multi-stage yields a slight gain on Erlangen-Prehistoric but degrades performance on Pforzheim Jewelry and Regensburg Scalalogy. We therefore treat multi-stage training as an empirical enhancement and report its effect comparatively rather than assuming consistent gains.

\subsection{Single-Dataset vs.\ Multi-Dataset Fine-tuning}
\label{sec:single_multi}

In addition to dataset-specific fine-tuning (one checkpoint per dataset), we fine-tune each of the selected top-3 models once on the union of all training sets (Erlangen-Prehistoric + Pforzheim Jewelry + Regensburg Scalalogy) using the same standard finetune training recipe. We then evaluate this multi-dataset checkpoint on each dataset's test set and compare it against the corresponding single-dataset checkpoint fine-tuned only on that dataset. This comparison isolates whether multi-dataset training improves generalization or introduces negative transfer across distinct schemas and text modalities.

\begin{table}[tbp]
\caption{Single-dataset vs.\ multi-dataset fine-tuning for the selected top-3 models. ``Single'' denotes a separate checkpoint fine-tuned per dataset; ``Multi'' denotes one checkpoint fine-tuned on the union of all training data and evaluated per dataset test set. All metrics are reported as percentages. $\downarrow$: lower is better; $\uparrow$: higher is better.}
\label{tab:single_multi}
\centering
\smalltablefont
\begin{tabular}{lccccccccr}
\toprule
\multirow{2}{*}{\textbf{Model}} & \multicolumn{3}{c}{\textbf{Erlangen-Prehistoric}} & \multicolumn{3}{c}{\textbf{Pforzheim Jewelry}} & \multicolumn{3}{c}{\textbf{Regensburg Scalalogy}} \\
\cmidrule(lr){2-4}\cmidrule(lr){5-7}\cmidrule(lr){8-10}
& CER$\downarrow$ & ANLS*$\uparrow$ & mAP-F1$\uparrow$ & CER$\downarrow$ & ANLS*$\uparrow$ & mAP-F1$\uparrow$ & CER$\downarrow$ & ANLS*$\uparrow$ & mAP-F1$\uparrow$ \\
\midrule
Gemma-3-4B (Single) & 10.4 & 88.6 & 83.1 & 0.9 & 99.0 & 99.4 & 14.6 & 73.8 & 46.9 \\
Gemma-3-4B (Multi) & \phantom{0}8.7 & 90.6 & 87.9 & 0.3 & 99.4 & 99.7 & \phantom{0}9.1 & 81.5 & 68.0 \\
Phi-3.5-Vision (Single) & 17.9 & 86.5 & 82.2 & 0.3 & 99.4 & 99.5 & \phantom{0}6.3 & 90.1 & 72.0 \\
Phi-3.5-Vision (Multi) & 19.7 & 84.0 & 45.4 & 0.5 & 99.2 & 99.4 & \phantom{0}7.4 & 84.3 & 54.1 \\
Qwen2.5-VL-7B (Single) & \phantom{0}5.8 & 91.0 & 90.0 & 0.5 & 99.2 & 99.4 & \phantom{0}7.6 & 87.0 & 62.8 \\
Qwen2.5-VL-7B (Multi) & \phantom{0}7.1 & 89.3 & 91.9 & 1.6 & 98.2 & 99.1 & 24.9 & 65.2 & 41.4 \\
\bottomrule
\end{tabular}
\end{table}

Overall, the effect of multi-dataset fine-tuning is clearly model-dependent rather than a uniform trade-off. For Gemma-3-4B, the multi-dataset checkpoint consistently outperforms the corresponding single-dataset checkpoints on all three datasets, improving both \ac{CER} and \ac{ANLS}$^\ast$ (Table~\ref{tab:single_multi}). In contrast, Qwen2.5-VL-7B shows the opposite behavior: the multi-dataset checkpoint generally degrades relative to single-dataset fine-tuning, with worse \ac{CER} and \ac{ANLS}$^\ast$ on all three datasets and the strongest drop on Regensburg Scalalogy (\ac{CER} 7.6\,\% $\rightarrow$ 24.90\,\%, \ac{ANLS}$^\ast$ 87.0\,\% $\rightarrow$ 65.2\,\%). 
While \ac{mAP-F1} slightly improves on Erlangen-Prehistoric (90.0\,\%$\rightarrow$ 91.9\,\%), it decreases on Pforzheim Jewelry (99.4\,\% $\rightarrow$ 99.1\,\%) and Regensburg Scalalogy (62.8\,\% $\rightarrow$ 41.4\,\%). 
Phi-3.5-Vision also degrades under multi-dataset fine-tuning on all three datasets, but the effect is markedly smaller than for Qwen, especially on Pforzheim Jewelry (\ac{CER} 0.30\,\% $\rightarrow$ 0.50\,\%, \ac{ANLS}$^\ast$ 99.4\,\% $\rightarrow$ 99.2\,\%). 
We therefore do not observe a single general rule for multi-dataset fine-tuning in this benchmark; instead, it can either improve cross-dataset robustness (Gemma) or introduce negative transfer of varying severity (Phi, Qwen) depending on the model.

\paragraph{Qualitative example.}
\Cref{fig:qualitative_gemma} illustrates a representative Erlangen-Prehistoric example for Gemma-3-4B, where fine-tuning improves field binding for inventory number and measurements, while some handwritten text and field-boundary errors remain.

\begin{figure}[!t]
\centering
\begin{minipage}[t]{0.48\textwidth}
\textbf{Zero-shot prediction}
\begin{lstlisting}[language=json,basicstyle=\ttfamily\tiny,aboveskip=1pt,belowskip=1pt]
{
 "Überschrift": "Anthropologisch-prähistorifche ...",
 "Inventarnummer": "J.-Nr. E024",
 "Maße": {"L": "", "B": "", "D": ""},
 "Objektbezeichnung": "Primitiver Kratzer",
 "Fundort": "Spay, Belgien",
 "Fundzeit": "Aumente Helin",
 "Beschreibungstext": "Flacher Stein ... [many OCR errors]"
}
\end{lstlisting}
\end{minipage}
\hfill
\begin{minipage}[t]{0.48\textwidth}
\textbf{Fine-tuned prediction}
\begin{lstlisting}[language=json,basicstyle=\ttfamily\tiny,aboveskip=1pt,belowskip=1pt]
{
 "Überschrift": "Anthropologisch=prähistorische ...",
 "Inventarnummer": "Eo 24",
 "Maße": {"L": "6,25", "B": "3,95", "D": "1,7"},
 "Objektbezeichnung": "Primitiver Kratzer ...",
 "Fundort": "Spy, Belgien Ausbeute Helin",
 "Fundzeit": "",
 "Beschreibungstext": "Flacher Stein ... [partially correct]"
}
\end{lstlisting}
\end{minipage}
\caption{Qualitative comparison for Gemma-3-4B on an Erlangen-Prehistoric sample. Fine-tuning corrects the inventory number and measurements and reduces sample-level CER from 26.8\,\% to 13.9\,\%, while longer handwritten text and \texttt{Fundort}/\texttt{Fundzeit} separation remain challenging.}
\label{fig:qualitative_gemma}
\vspace{-0.8em}
\end{figure}

\section{Conclusion}
\label{sec:conclusion}

We benchmarked eight lightweight \acp{VLM} for document \ac{OCR} with schema-guided \ac{JSON} extraction across three archival datasets covering handwritten inventory books, mostly printed jewelry records, and structurally complex staircase forms. Overall, instruction-tuned general-purpose \acp{VLM} (Qwen2.5-VL, Phi-3.5-Vision, and Gemma-3) are the strongest candidates, outperforming several OCR-specialized alternatives in structured generation. This indicates that reliable extraction depends not only on text recognition, but also on instruction following, field binding, and stable schema-conformant decoding.

Few-shot prompting is not consistently beneficial: it can help on cleaner printed records such as Pforzheim Jewelry, but remains mixed or negative on handwritten and structurally complex datasets, where longer prompts can increase instruction complexity. Fine-tuning is the most effective adaptation strategy, with full fine-tuning for sub-1B models and \ac{QLoRA} for larger models enabling strong local performance on a single consumer \ac{GPU}. Among the additional studies, hyperparameter optimization provides the most reliable modest gains, particularly on the more difficult Regensburg Scalalogy dataset, whereas image preprocessing and multi-stage training are strongly model- and dataset-dependent and can also degrade structured extraction.

The single- versus multi-dataset experiments further show that joint training is not uniformly beneficial: Gemma-3-4B improves across all three collections, suggesting better cross-schema robustness, whereas Phi-3.5-Vision and especially Qwen2.5-VL-7B show signs of negative transfer. This highlights that deployment decisions should consider whether an institution needs one general checkpoint across collections or separate dataset-specific checkpoints for maximum accuracy.

Our study has two main limitations: we evaluate hyperparameter optimization, preprocessing, and multi-stage training independently, and the lightweight \ac{VLM} landscape evolves rapidly. Thus, our findings should be read as guidance for the evaluated model families rather than an exhaustive statement about all emerging OCR/document models. Future work should explore constrained decoding for valid \ac{JSON}, layout-aware field binding, handwriting-specific adaptation, broader multilingual datasets, and controlled comparisons of full fine-tuning versus \ac{QLoRA} under matched compute budgets.

\subsubsection*{Disclosure of Interests}
The authors declare no competing interests.

\subsubsection*{Acknowledgments}
This work was supported by the German Federal Ministry of Research, Technology and Space (BMFTR) within the project Sammlungen, Objekte, Datenkompetenzen (SODa), which aims to establish a data competence center for university collections.
The authors also gratefully acknowledge the scientific support and HPC resources provided by the Erlangen National High Performance Computing Center (NHR@FAU) of the Friedrich-Alexander-Universität Erlangen-Nürnberg (FAU) under the NHR project b268dc. NHR funding is provided by federal and Bavarian state authorities.
We extend our gratitude to Tabea Schmid and Katharina Wittemann from the Faculty of Design at HS Pforzheim, Sophie Schlosser from the Friedrich-Mielke Institute for Scalalogy at the OTH Regensburg, as well as Doris Mischka and Thorsten Uthmeier from the Institut für Vor- und Frühgeschichte at the FAU Erlangen. 
Their domain expertise, constructive criticism, and support were invaluable for our experiments.

\bibliographystyle{splncs04} 
\bibliography{refs} 

\end{document}